\documentclass[conference]{IEEEtran}
\usepackage{cite}
\usepackage{multirow}
\usepackage{pdflscape}
\usepackage{colortbl}
\usepackage{array}
\usepackage{tikz}
\usepackage{subcaption}
\usepackage{graphicx}
\usepackage{float}
\usepackage{booktabs}
\usepackage{mathrsfs}
\usepackage{amsmath, amssymb, amsfonts}
\usepackage[ruled,vlined]{algorithm2e}
\usepackage{orcidlink}
\usepackage{tikz}
\usepackage[normalem]{ulem}
\useunder{\uline}{\ul}{}
\def\BibTeX{{\rm B\kern-.05em{\sc i\kern-.025em b}\kern-.08em
    T\kern-.1667em\lower.7ex\hbox{E}\kern-.125emX}}

\makeatletter
\renewcommand\@makefnmark{} 
\makeatother

\begin{document}

\title{Unlearning Backdoor Attacks through Gradient-Based Model Pruning} 

\author{\IEEEauthorblockN{Kealan Dunnett\textsuperscript{1,2,\dag}\thanks{\textsuperscript{\dag}Corresponding Author: kealan.dunnett@hdr.qut.edu.au}\orcidlink{0000-0002-0010-1499}, Reza Arablouei\textsuperscript{2}\orcidlink{0000-0002-6932-2900}, Dimity Miller\textsuperscript{1}\orcidlink{0000-0001-6312-8325}, Volkan Dedeoglu\textsuperscript{1,2}\orcidlink{0000-0002-2567-2423}, Raja Jurdak\textsuperscript{1}\orcidlink{0000-0001-7517-0782}}

\IEEEauthorblockA{\textsuperscript{1}\textit{Queensland University of Technology}, \textsuperscript{2}\textit{CSIRO's Data61}}}

\IEEEaftertitletext{\footnotetext[\dag]{Corresponding Author: kealan.dunnett@hdr.qut.edu.au}}

\maketitle

\begin{abstract}
In the era of increasing concerns over cybersecurity threats, defending against backdoor attacks is paramount in ensuring the integrity and reliability of machine learning models. However, many existing approaches require substantial amounts of data for effective mitigation, posing significant challenges in practical deployment. 
To address this, we propose a novel approach to counter backdoor attacks by treating their mitigation as an unlearning task. We tackle this challenge through a targeted model pruning strategy, leveraging unlearning loss gradients to identify and eliminate backdoor elements within the model. Built on solid theoretical insights, our approach offers simplicity and effectiveness, rendering it well-suited for scenarios with limited data availability. Our methodology includes formulating a suitable unlearning loss and devising a model-pruning technique tailored for convolutional neural networks. Comprehensive evaluations demonstrate the efficacy of our proposed approach compared to state-of-the-art approaches, particularly in realistic data settings.
\end{abstract}



\section{Introduction}\label{introduction}

\footnotetext{\textsuperscript{\dag}Corresponding author: Kealan Dunnett (kealan.dunnett@hdr.qut.edu.au)}

Recently, deep learning has witnessed remarkable advancements, leading to its widespread adoption across various industries and academic fields. For example, sectors such as healthcare, education, automobiles, and logistics have rapidly embraced deep learning technologies~\cite{pouyanfar2018survey}. However, alongside its positive impact, deep learning also faces significant adversarial threats that pose challenges to its practical implementation~\cite{xue2020machine}.

Backdoor attacks represent a critical adversarial threat in classification-based machine learning tasks. Initially demonstrated by \cite{gu2019badnets}, these attacks seek to compromise the integrity of model decisions by altering predictions when presented with inputs containing a specific trigger. The primary objective of backdoor attacks is to embed this behavior into the model without undermining its ability to classify clean inputs (i.e., inputs lacking the trigger). Consequently, a backdoored model distinguishes between inputs with the trigger and those without it during classification. Given the complexity of modern deep neural network models, there is typically no overt indication that such embedded backdoor behavior exists.

In real-world scenarios, backdoor attacks pose a significant threat in applications where classification model outputs inform automated decision-making processes. A notable example is traffic sign classification, particularly with the increasing adoption of driver assistance systems. Misclassification of critical traffic signs with these systems can have catastrophic consequences~\cite{pouyanfar2018survey}. The risk of backdoor attacks is especially pertinent when model training is outsourced to a third party (e.g., through cloud services or federated learning) or when transfer learning is employed~\cite{liu2019abs}, as this provides opportunities for adversaries to manipulate the model's training data and or procedures to inject backdoor tasks.

In response to the threats posed by backdoor attacks, various methods have been proposed for detecting the presence of backdoors within models as well as extracting the respective backdoor triggers using clean inputs. For instance, \cite{liu2019abs} introduces an optimisation-based method to discover a trigger pattern that can transition a set of images to a target class with minimal input perturbation. As well as backdoor discovery methods, numerous backdoor mitigation approaches have also been proposed in the literature \cite{wu2021adversarial,wang2019neural,zhao2020bridging}. These approaches aim to remove backdoor behavior from a model using clean or backdoor inputs with minimal impact on the original classification objective. To achieve this, several existing works leverage model pruning, a technique traditionally used to sparsify overparameterized models, which has proven effective for backdoor mitigation. 

Existing backdoor attack mitigation approaches based on model pruning mainly rely on analyzing neuron activation values~\cite{liu2018fine,zheng2022data,li2020neural}, sensitivity~\cite{wu2021adversarial}, or reconstruction~\cite{li2023reconstructive}. While these approaches have shown effectiveness within certain settings, their performance within data-limited settings, common in real-world applications, remains unclear. Moreover, many of these approaches require the defender to carefully select the values of one or more hyperparameters, as in~\cite{zheng2022data, zheng2022data, wu2021adversarial}. While we acknowledge that achieving a hyperparameter-free pruning approach may be infeasible, we advocate for the development of approaches that only mandate the selection of few intuitive hyperparameters, minimizing the need for extensive hyperparameter tuning. A notable example is~\cite{liu2018fine}, where the defender is only required to specify an acceptable accuracy reduction threshold.

\textbf{Contributions:} We propose a new approach for backdoor attack mitigation by employing gradient-based model pruning to unlearn backdoor behavior. By conceptualizing backdoor mitigation as an unlearning problem, our approach harnesses information from unlearning loss gradients to effectively eliminate backdoors through model pruning. This sets our approach apart from existing model-pruning-based approaches to backdoor mitigation, which typically rely on information available through neuron activation values, sensitivity, or reconstruction. 
Moreover, we incorporate an effective fine-tuning procedure that utilizes both clean and backdoor data to address any performance degradation resulting from model pruning.  
A key advantage of the proposed approach is its user-friendly nature, as defenders are only required to adjust a few design parameters with minimal or no tuning. The selection of these parameters is tailored to defenders' specific needs and does not demand intricate tuning, unlike the hyperparameters utilized in many existing approaches.
We conduct comprehensive evaluations of our approach against state-of-the-art approaches, demonstrating its effectiveness across various limited data scenarios. Our evaluations also offer fresh insights into the performance of existing state-of-the-art methods in constrained data scenarios.

\section{Related Work} \label{related-work}

In this section, we provide a summary of prominent backdoor attack and mitigation works. For each topic, the range of strategies used by these works is highlighted. 

\subsection{Backdoor Attacks} \label{related-attacks}

The BadNets attack, introduced in \cite{gu2019badnets}, stands as a seminal example of a backdoor attack. This work demonstrated how manipulating a subset of training data by inserting a colored square can effectively embed a backdoor task into a model. Building upon this foundation, subsequent studies have proposed diverse attack methods of varying complexity. For instance, the blended attack, outlined in \cite{chen2017targeted}, modifies training instances by incorporating a trigger image (e.g., a picture of Hello Kitty) with a blending ratio that controls its transparency. Furthermore, \cite{barni2019new} proposes a backdoor attack that introduces an extra feature to images in the training set associated with the target class (e.g., a sinusoidal signal). Following training, \cite{barni2019new} demonstrates that incorporating this feature into images from other classes causes them to be misclassified as the target class. 

More recently, several optimisation-based backdoor attacks have been proposed. For example, \cite{li2021invisible} presents a method employing an encoder-decoder network to generate sample-specific triggers for training a backdoored model. Moreover, \cite{doan2021lira} presents a backdoor attack framework that jointly optimizes the trigger pattern and learns a backdoored model. This approach yields a targeted trigger representation that fulfills the backdoor objective while remaining stealthy. Beyond trigger-based methods, alternative relabeling approaches have emerged. For example, \cite{zhao2020bridging} proposes an all-to-all backdoor attack, wherein any input containing the backdoor trigger is mapped to a class one step ahead in the cyclical order, i.e., $y + 1 \mod n$ where $y$ is the true class label and $n$ the number of classes. However, our work aligns with the prevailing trend in the literature by focusing solely on the targeted attack setting, where the trigger steers the classification towards a static target class $t$.

\subsection{Backdoor Mitigation}

Model pruning and fine-tuning using clean training data, as proposed by \cite{liu2018fine}, represent the initial steps towards removing backdoors from trained models. Subsequent works, such as \cite{liu2018fine,wang2019neural,wu2021adversarial,zheng2022data}, have further developed model pruning. These works develop a range of approaches based on the observation that clean and backdoor inputs produce different activation values throughout the model. Subsequently, by inferring these differences using a variety of techniques, these proposals successfully identify and mitigate backdoor elements, which are components of the model that contribute the most to the backdoor task. Notably, \cite{zheng2022data} proposes a data-free pruning approach, eliminating the need for clean or backdoor data during model pruning.

In addition, several fine-tuning and model regularisation approaches have also been presented in the literature. For example, \cite{li2020neural} introduces an attention-based mechanism for backdoor removal. Utilizing a fine-tuned teacher network, \cite{li2020neural} employs a layer-based knowledge distillation technique to eliminate the backdoor from the model. Moreover, works such as \cite{liu2022backdoor,chen2022effective} propose unlearning-like approaches that remove backdoors through fine-tuning alone. However, these approaches primarily focus on model fine-tuning rather than model pruning.

\section{Preliminaries} \label{preliminaries}

\subsection{Neural Networks} \label{prelim-neural-networks}

A neural network is a parameterized function $f(x,\theta)\rightarrow y$ that maps an input $x\in\mathbb{R}^{n}$ to output $y\in\mathbb{R}^{m}$ given a set of parameters $\theta$. In an $m$-class classification scenario, the entries of $y$ represent the likelihood of $x$ belonging to each class. Using a training dataset ($D_t: \{x_{i}, y_{i}\}_{i=1}^{Z}$), we can find the optimal set of parameters $\theta^*$ by minimizing the aggregate loss function, $\mathscr{L}(\hat{y},y)$, which quantifies the difference between each predicted value $\hat{y}_i=f(x_i,\theta)$ and its corresponding actual value $y_i$. Therefore, we have
\begin{equation}\label{optimisation}
    \theta^* = \arg\min_{\theta} \sum_{i=1}^{Z} \mathscr{L}(f(x_{i}, \theta), y_{i}).
\end{equation}
This optimization is typically performed using the stochastic gradient descent algorithm or one of its variants.

\subsection{Backdoor Threat Model} \label{threat-model}

In a targeted attack scenario, the adversary embeds a backdoor task into the model, causing inputs containing a trigger pattern $p$ to be classified as the target class $t$. This manipulation involves training the model using two datasets: the clean dataset $D_c=\{x_{i}, y_{i}\}_{i=1}^{Z_c}$, comprising original inputs and their labels, and the backdoor dataset $D_b=\{\breve{x}_{i}, t\}_{i=1}^{Z_b}$ generated by the adversary using the trigger pattern $p$. 
Note that, in certain attack scenarios, such as \cite{doan2021lira}, $p$ may vary with the input, rendering it dynamic.   
Using $D_c$ and $D_b$, the adversary aims to determine a set of model parameters $\theta'$ capable of effectively classifying inputs from both sets. In this optimization process, the adversary employs a poisoning ratio to specify the proportion of backdoor to clean inputs.

\subsection{Main Assumptions} \label{threat-model-assumptions}

We adopt common assumptions regarding the capabilities of adversaries and defenders. Concerning the attacker, we consider a range of attacks documented in the literature, without imposing any additional constraint beyond those stated in the corresponding original papers. For instance, in the most extreme case, as depicted in the BPP attack by \cite{wang2022bppattack}, it is assumed that the attacker possesses complete control over the training process. However, to streamline our evaluations, we focus solely on the targeted attack scenarios, thereby reducing the complexity of the analysis. 

We assume the defender has access to a limited set of correctly labeled clean images (i.e., images without the backdoor pattern). In practice, the number of clean samples accessible to the defender can be severely constrained. 
In addition, we assume that the defender can synthesize backdoor inputs using any relevant state-of-the-art synthesis approach, such as those proposed in \cite{liu2022backdoor,liu2019abs,sun2023single}. Thus, the defender has access to a backdoor variant for each clean image, incorporating the adversary's trigger. While acknowledging that this assumption limits our approach's applicability to scenarios where faithful synthesis of backdoor inputs is not feasible, we aim to address this limitation in future work.

\section{Proposed Approach} \label{proposed-approach}

In this section, we begin with describing our conceptualization of backdoor unlearning. We then describe the proposed approach, which consists of two key steps. The first step involves iteratively pruning the backdoored model using the gradient of unlearning loss. In the second step, we fine-tune the pruned model utilizing available clean and backdoor data to alleviate any adverse affect of model pruning.

\subsection{Backdoor Unlearning} \label{backdoor-unlearning}

For a backdoored model to effectively fulfil the adversarial objective outlined in section~\ref{threat-model}, it must accurately classify inputs associated with both the main and backdoor tasks. The main task is to correctly classify the clean inputs from $D_c$ and the backdoor task is to correctly classify the backdoor inputs from $D_b$. Therefore, we conceptualize the challenge of backdoor mitigation as an unlearning problem, where the goal is to unlearn the backdoor task while preserving the efficacy of the main task. As such, backdoor unlearning aims to achieve three primary objectives: (i) Eliminate the model's association of the given trigger pattern $p$ with the target class $t$. (ii) Rectify the model's interpretation of $p$ to ensure that inputs from $D_b$ are classified as their original labels. (iii) Ensure that the model's correct classification of inputs from $D_c$ is maintained.

While the concept of unlearning has been employed in prior works such as \cite{liu2022backdoor,wang2019neural,chen2022effective} to mitigate backdoor attacks, our unlearning approach differs significantly in that backdoor unlearning is achieved through gradient-informed model pruning. In existing works, unlearning typically involves model fine-tuning using a modified loss function. For example, in \cite{liu2022backdoor}, unlearning is achieved through model retraining using a loss function comprising three components, one of which is the negative cross-entropy loss of the backdoor inputs given the backdoor label and thus induces unlearning through gradient ascent.
Unlike previous works, we introduce a backdoor unlearning approach that is implemented through gradient-based model pruning. Our backdoor unlearning loss function is the aggregate cross-entropy loss for the backdoor inputs with their corresponding correct labels. We express this loss as
\begin{equation} \label{unlearning-loss}
    \mathcal{L}=\sum_{i=1}^{Z_b} \mathscr{L}_{\text{CE}}(f'(\breve{x}_i, \theta'), y_{i})
\end{equation}
Where $f'$ and $\theta'$ denote the backdoored model and its parameters, respectively.
Hence, we compute the loss gradient with respect to $\theta'$ as $\nabla_{\theta'}=\frac{\partial \mathcal{L}}{\partial \theta'}$.

Our approach does not explicitly aim to minimize the above unlearning loss. Instead, we utilize its gradient information to gauge the relative contribution of each parameter subset to the backdoor behavior. Larger gradients in the unlearning loss suggest a stronger influence on misclassifying $\breve{x}_i$ as $t$ instead of its true label $y_{i}$. However, unlike traditional model (un)learning, which adjusts these parameters, we opt to prune them entirely. This decision is rooted in the understanding that such parameters are typically manipulated by the backdoor attack in a way that adjusting them through stochastic gradient descent based on limited data is ineffective in mitigating the backdoor attack. Pruning these parameters removes their influence on the model's behavior, thereby effectively mitigating the backdoor effect. We elaborate the details of this pruning strategy in the next section.

\subsection{Gradient-based Pruning} \label{pa-pruning}

\begin{figure*}[th]
    \centering
    \begin{subfigure}[b]{1\textwidth}
        \centering
        \includegraphics[width=\textwidth]{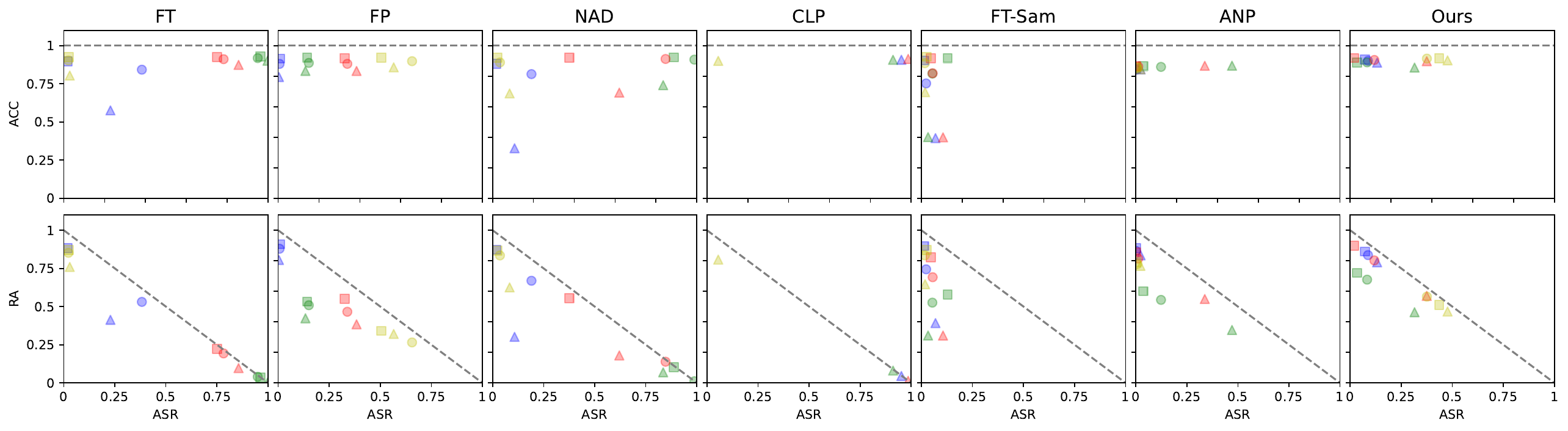}
        \caption{PreactResNet-18}
        \label{fig:cifar-preactresnet}
    \end{subfigure}
    \begin{subfigure}[b]{1\textwidth}
        \centering
        \includegraphics[width=\textwidth]{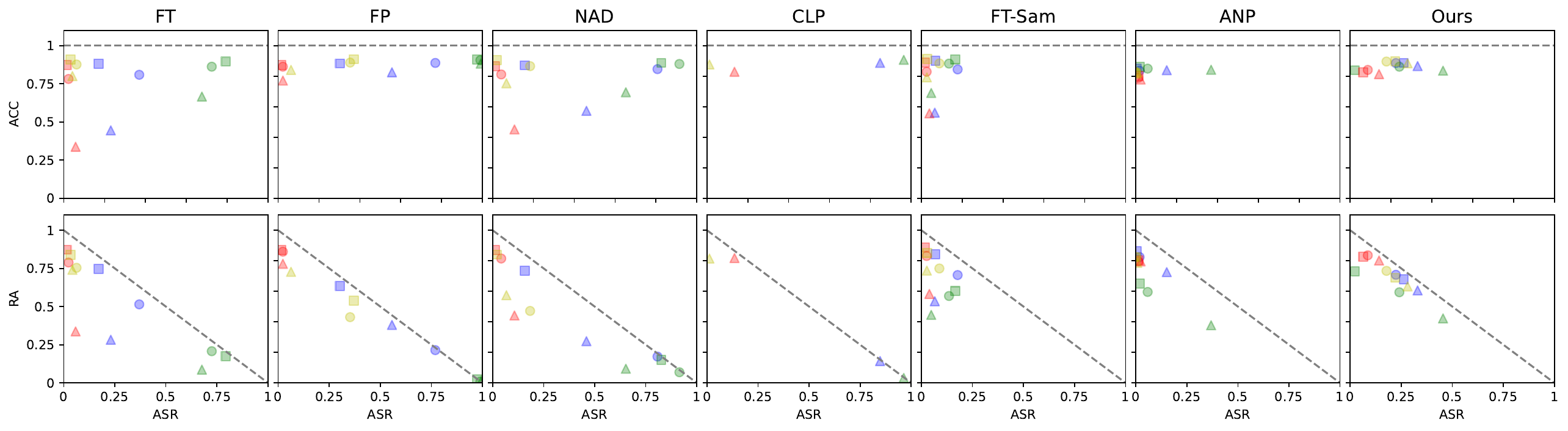}
        \caption{VGG-19+BN}
        \label{fig:cifar-vgg19}
    \end{subfigure}
    \begin{subfigure}[c]{1\textwidth}
        \centering
        \includegraphics[width=0.95\textwidth]{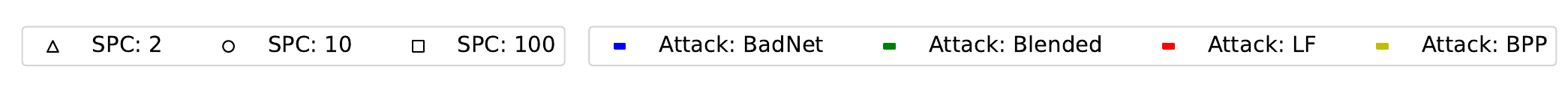}
    \end{subfigure}
\caption{The scatter plots of ACC and RA versus ASR for all considered approaches on CIFAR-10 across different attacks, SPC settings, and models.}  
\label{fig:cifar-results}
\end{figure*}

Building on the described notion of backdoor unlearning, we propose a gradient-based pruning technique tailored for convolutional layers. For each 2D convolutional filter $i$ at layer $l$ with parameters $\theta'_{l,i}$, we compute the mean absolute gradient as
\begin{equation} \label{MAGM}
    \xi_{l,i} = \frac{\|\nabla_{\theta'_{l,i}}\|_{1}}{\ell(\nabla_{\theta'_{l,i}})}
\end{equation}
where $\|\cdot\|_1$ and $\ell(\cdot)$ denote the L1 norm and the number of entries, respectively. 
Given the nature of the unlearning loss presented in section~\ref{backdoor-unlearning}, $\xi_{l,i}$ represents the relative contribution of filter $i$ at layer $l$ to misclassification of backdoor inputs.   
After calculating the $\xi_{l,i}$ values for all filters across all layers, we identify the filter with the highest $\xi_{l,i}$ value for pruning. 
To prune a filter, we set its weights and bias to zero. Following each pruning round, we evaluate the unlearning loss and main task accuracy for the pruned model using a separate validation dataset that is not used for calculating the filter $\xi_{l,i}$ values. We iterate the pruning process until the main task accuracy falls below a predefined threshold $\alpha$ or the unlearning loss fails to improve for $P_{p}$ consecutive rounds. 


\subsection{Fine-tuning} \label{pa-tuning}

As demonstrated in \cite{liu2018fine}, model pruning aimed at countering backdoor attacks can often lead to a degradation in the model's accuracy when classifying clean inputs. To recover the lost performance resulting from model pruning, we adopt a modified version of the fine-tuning approach proposed in~\cite{wang2019neural}. Additionally, since our model pruning technique targets convolutional layers, fine-tuning facilitates the removal of backdoor elements present in other unpruned layers. This is particularly beneficial when backdoor elements are present in the dense layers.

To perform fine-tuning, we propose to utilize all available clean and backdoor data that, as discussed in section~\ref{threat-model-assumptions}, typically constitute a small subset of the original training dataset. This is in contrast with previous approaches such as \cite{wang2019neural} that use a portion of the backdoor data during fine-tuning. Additionally, each backdoor datum is labeled with its correct (non-backdoor) label. The fine-tuning process, which is essentially model re-training, continues until the loss fails to improve for a specified number of epochs/iterations $P_{t}$. This ensures that the performance of the main task is maintained throughout fine-tuning. Similar to our pruning approach discussed in section~\ref{pa-pruning}, we evaluate the loss using a separate validation set that is not used for fine-tuning.

\section{Evaluation}

\begin{figure*}[t]
    \centering
    \begin{subfigure}[b]{0.49\textwidth}
        \centering
        \includegraphics[width=\textwidth]{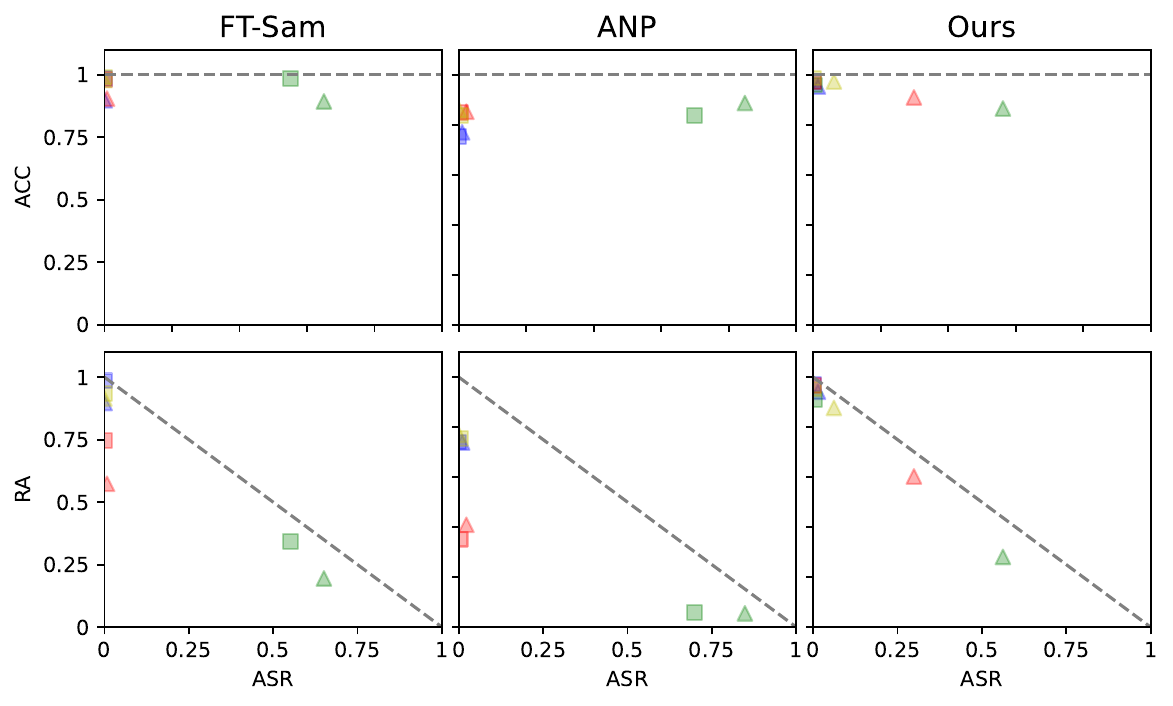}
        \caption{PreactResNet-18}
        \label{fig:gtsrb-preactresnet}
    \end{subfigure}
    \begin{subfigure}[b]{0.49\textwidth}
        \centering
        \includegraphics[width=\textwidth]{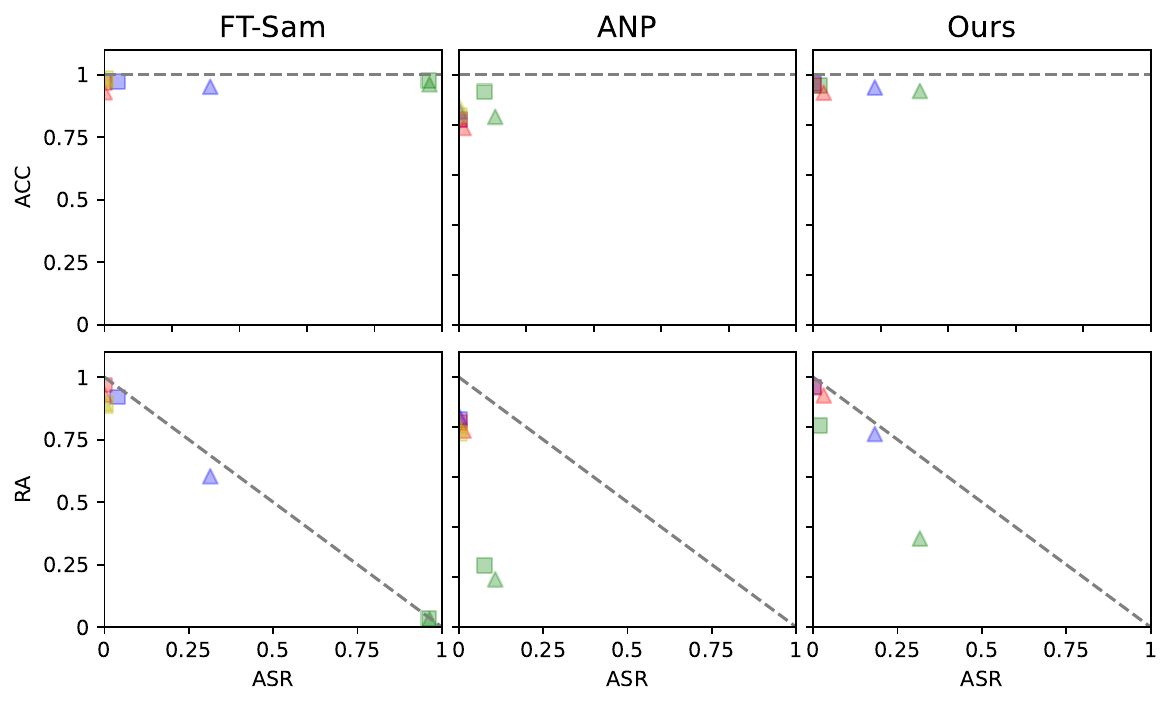}
        \caption{VGG-19+BN}
        \label{fig:gtsrb-vgg19}
    \end{subfigure}
    \begin{subfigure}[b]{0.49\textwidth}
        \centering
        \includegraphics[width=\textwidth]{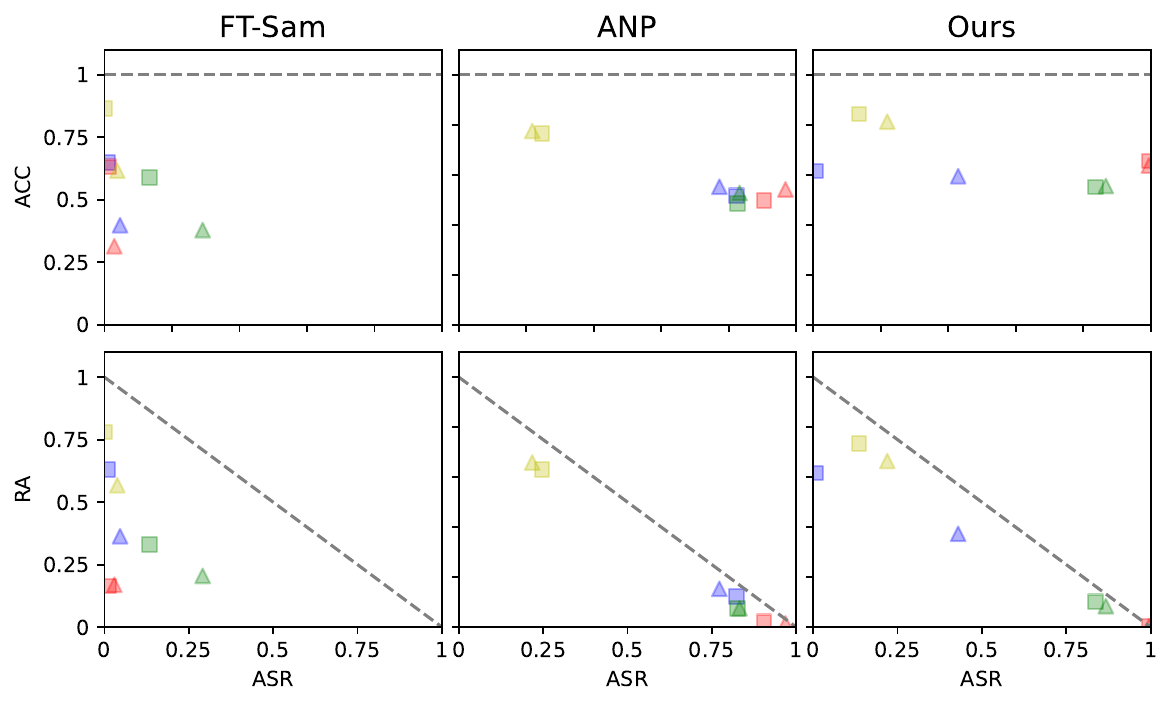}
        \caption{EfficientNet B3}
        \label{fig:gtsrb-efficientnet}
    \end{subfigure}
    \begin{subfigure}[b]{0.49\textwidth}
        \centering
        \includegraphics[width=\textwidth]{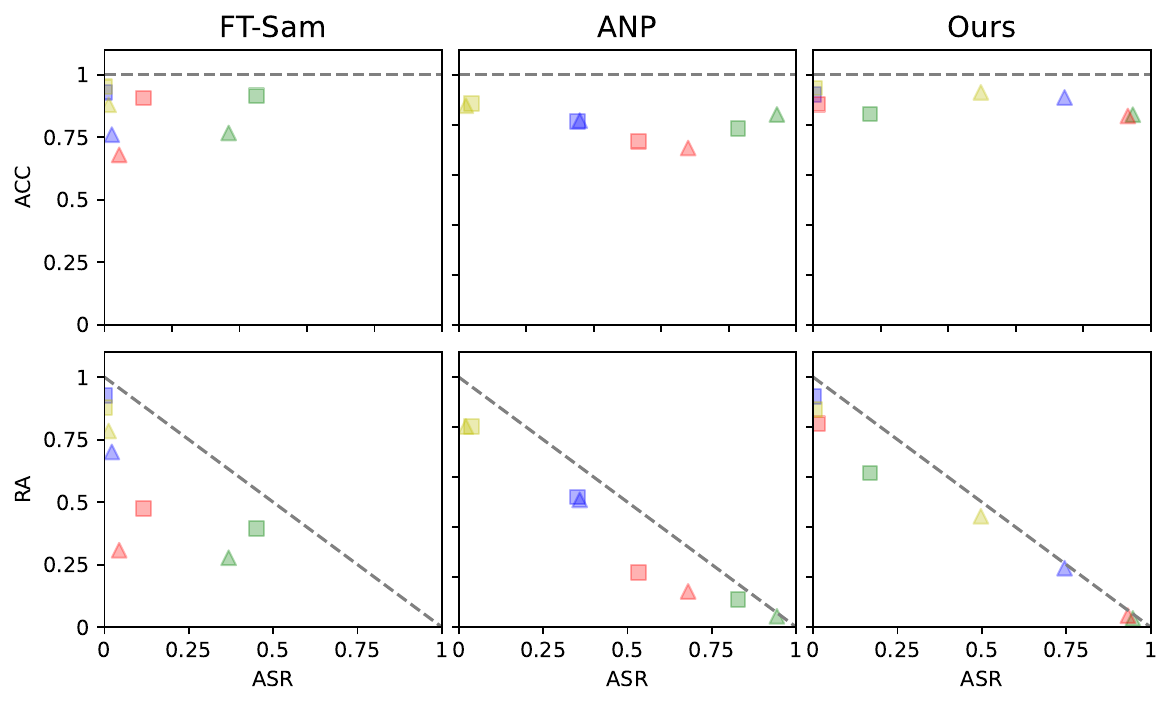}
        \caption{MobileNet V3 Large}
        \label{fig:gtsrb-mobilenet}
    \end{subfigure}
    \begin{subfigure}[c]{1\textwidth}
        \centering
        \includegraphics[width=0.9\textwidth]{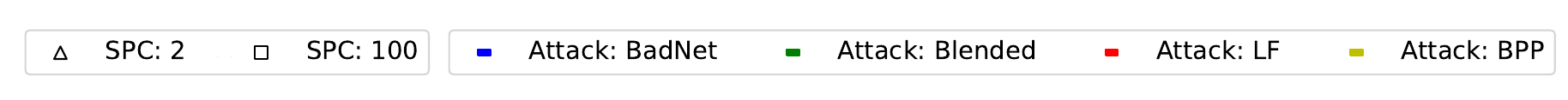}
    \end{subfigure}

\caption{The scatter plots of ACC and RA versus ASR for FT-SAM, ANP, and the proposed approach on GTSRB across different attacks, SPC settings, and models.}
\label{fig:gtsrb-results}
\end{figure*}

We evaluate the performance of our proposed approach against five existing backdoor mitigation approaches across four types of attacks and various settings. To this end, we utilize the BackdoorBench tool presented in~\cite{wu2022backdoorbench}. The code to reproduce the results presented in this paper is available online\footnote{\url{https://github.com/WhoDunnett/Grad-Prune/tree/main}}.

\subsection{Attack Configuration}

In our evaluations, we consider four backdoor attacks of BadNets \cite{gu2019badnets}, Blended \cite{chen2017targeted}, Low Frequency (LF) \cite{zeng2021rethinking} and Bit-Per-Pixel (BPP) \cite{wang2022bppattack}. According to~\cite{wu2022backdoorbench}, these attacks encompass a range of trigger characteristics utilized in state-of-the-art attacks. For each attack, we use the default configuration provided in~\cite{wu2022backdoorbench}. Notably, we only considered a 10\% poisoning setting and an all-to-one targeted attack, with the target class label being 0. We implement the attacks on the CIFAR-10~\cite{krizhevsky2009learning} and German Traffic Sign Recognition Benchmark (GTSRB)~\cite{Houben-IJCNN-2013} datasets, employing multiple model architectures. For CIFAR-10, we consider the PreactResNet-18 model and the VGG-19 model incorporating batch normalization (referred to as VGG-19+BN). For GTSRB, we examine the same models along with the EfficientNet B3 and MobileNet V3 Large models. We utilize all models with their default configurations provided in~\cite{wu2022backdoorbench}. For each configuration, the baseline results correspond to when no backdoor mitigation approach is applied.

\subsection{Defense Configuration}

We benchmark our proposed backdoor attack mitigation approach against several commonly used approaches, including Fine-Tuning~\cite{liu2018fine}, Fine-Pruning~\cite{liu2018fine} and NAD~\cite{li2021neural}. In addition, we include the state-of-the-art defences CLP~\cite{zheng2022data}, FT-SAM~\cite{zhu2023enhancing} and ANP~\cite{wu2021adversarial}. We test FT-SAM and ANP only on GTSRB, based on their performance on CIFAR-10. For each defence, we use the implementation provided in BackdoorBechmark~\cite{wu2022backdoorbench}, using default configurations. We do not perform hyperparameter optimization and rely on default settings reported in the relevant papers, since conducting such optimization is typically infeasible for defenders~\cite{wu2022backdoorbench}.

Unlike existing works, we evaluate the performance of each considered defensive approach across a range of limited data settings. Existing approaches usually use a percentage of training data, often 1\% or 10\%, to evaluate the efficacy of their proposals. However, using a proportion of training data can make comparisons across datasets challenging due to variations in dataset sizes. Moreover, accessing even 1\% of training data, especially in large datasets, may be unrealistic. To better reflect the data available to defenders in practical scenarios, we conduct our evaluations considering 2, 10, or 100 samples per class (SPC).
For each dataset, model architecture, and attack type, we test every backdoor mitigation approach five times, with each trial using a different subset of data. When validation data is required, such as for Fine-Pruning, ANP, and our approach, we use 10\% of the data for validation. For the 2 SPC cases, we use one sample for training (fine-tuning) and the other for validation. 

\subsection{Performance Measures}

We evaluate the performance of each approach across all settings using three accuracy measures: accuracy on clean data (ACC), attack success rate (ASR), and recovery accuracy (RA) \cite{wu2022backdoorbench}. ACC represents the classification accuracy on the clean test dataset devoid of any backdoor attack. Moreover, ASR denotes the accuracy on the test dataset containing the backdoor trigger and the backdoor target labels, while RA indicates the accuracy on the test dataset containing the backdoor trigger and the correct non-backdoor labels. Note that $\text{ASR} + \text{RA} \leq 1$. A successful defence is characterized by high values of ACC and RA and a low value of ASR.  

\subsection{Results}

In Fig.~\ref{fig:cifar-results}, we depict the performance of various approaches using scatter plots of ACC and RA versus ASR for the CIFAR-10 dataset, employing the PreactResNet-18 and VGG-19+BN models across different attacks and data settings. We provide the corresponding accuracy measure values used to generate these figures in Tables~\ref{tab:cifar-10-PreactResNet-18} and \ref{tab:cifar-10-VGG19+BN} as supplementary material. Except for the BPP attack, our approach exhibits competitive performance compared with the state-of-the-art approaches FT-SAM and ANP. In most cases, our approach significantly reduces ASR compared to the baseline with minimal impact on ACC. This decrease in ASR often coincides with a proportional rise in RA.
While our approach may show slightly less effectiveness in low-data settings (i.e., low SPC) compared to FT-SAM and ANP, overall, it demonstrates promising results across various scenarios.
The results in Fig.~\ref{fig:cifar-results} also indicate that the effectiveness of FT, FP, and NAD varies noticeably across the considered scenarios. More specifically, the effectiveness of these approaches is significantly impacted by the amount of available data.

In addition, Fig.~\ref{fig:cifar-results} shows that FT, FP, and NAD exhibit significant variability in their performance compared to FT-SAM, ANP, and our approach.
While CLP demonstrates good performance in specific scenarios, it proves ineffective in several cases, indicating that its underpinning assumptions may not hold universally across all model architectures, especially those not examined in the original paper~\cite{zheng2022data}.
Overall, FT-SAM appears to mitigate the backdoor attacks most successfully among the tested approaches. However, its RA exhibits greater variance compared to ANP and our approach. In addition, RA of FT-SAM is limited in low-data settings, suggesting that while it may mitigate the backdoor attack, it does not consistently restore the correct classification of backdoor images.

When compared to the activation-based pruning approaches FP and CLP, our approach is generally significantly more effective. While FP and CLP may exhibit state-of-the-art performance in certain cases, their overall effectiveness varies greatly. Therefore, we posit that the backdoor unlearning loss gradient can capture the essential information required for backdoor removal more adeptly compared to activation values.

In Fig.~\ref{fig:gtsrb-results}, we illustrate the results for the GTSRB dataset, utilizing the PreactResNet-18, VGG-19+BN, EfficientNet B3, and MobileNet V3 Large models. Here, we only consider FT-SAM, ANP, and our approach as they perform considerably better over CIFAR-10. Results for PreactResNet-18 and VGG-19+BN closely align with those reported for CIFAR-10, showing consistent and effective backdoor attack mitigation across various settings. Notably, our approach exhibits less variability compared to FT-SAM and ANP with both models. However, for EfficientNet B3 and MobileNet V3 Large, greater variance is observed overall. Specifically, all three approaches face challenges in mitigating the backdoor attacks when employing these models. With MobileNet V3 Large, our approach demonstrates superior robustness compared to FT-SAM and ANP in the $\text{SPC}=100$ setting. While our evaluation results do not conclusively establish gradient-informed model pruning as yielding superior performance compared to FT-SAM and ANP, its competitive performance underscores the utility of unlearning loss gradients. Moreover, our findings corroborate the idea of reframing backdoor attack mitigation as an unlearning problem addressed through model pruning.

\section{Conclusion}

We investigated the efficacy of gradient-informed model pruning for mitigating backdoor attacks. By casting backdoor attack mitigation as an unlearning task addressed via model pruning, we leveraged gradients of an aptly-devised unlearning loss to remove backdoor elements. Our proposed approach offers an intuitive yet theoretically well-grounded solution to backdoor mitigation. Moreover, it demonstrates robust performance across various scenarios, surpassing or rivalling state-of-the-art approaches. Encouraged by our findings, we advocate for further exploration of gradient-based techniques in future research. The prospect of eliminating the need for synthesizing backdoor data is particularly promising. Furthermore, our benchmarking highlights the importance of addressing challenges in low-data settings and considering a diverse range of model architectures in future defense strategies.

\section{Acknowledgement}
Computational resources and services used in this work were provided by the eResearch Office, Queensland University of Technology, Brisbane, Australia.

\bibliographystyle{splncs04}
\bibliography{egbib}

\begin{landscape}
\section{Supplementary Material}
\begin{table}[h]
\caption{The performance of the considered approaches on CIFAR-10 using PreactResNet-18. The mean and standard deviation of the performance measures over five independent runs are presented. In each setting, the best and second-best results are highlighted in bold and underlined, respectively.}
\scalebox{0.95}{
\begin{tabular}{c|c|ccc|ccc|ccc|ccc}
\hline
\multicolumn{2}{c}{Attack}                           & \multicolumn{3}{c}{BadNet}                                     & \multicolumn{3}{c}{Blended}                                    & \multicolumn{3}{c}{BPP}                                        & \multicolumn{3}{c}{LF}                                          \\ \hline 
SPC                              & Defense           & ACC                 & ASR                & RA                  & ACC                 & ASR                & RA                  & ACC                 & ASR                & RA                  & ACC                 & ASR                & RA                   \\ \hline
\rowcolor[HTML]{E8E8E8} 
\multicolumn{2}{c}{\cellcolor[HTML]{E8E8E8}Baseline} & \textit{91.82}      & \textit{94.76}     & \textit{5}          & \textit{93.61}      & \textit{99.71}     & \textit{0.29}       & \textit{91.46}      & \textit{99.93}     & \textit{0.07}       & \textit{93.2}       & \textit{99}        & \textit{0.93}        \\ \hline
                                 & FT                & 57.53±24.27         & 22.86±30.4         & 41.31±20.27         & {\ul 90.11±4.0}     & 99.75±0.51         & 0.19±0.44           & 80.35±2.83          & 2.93±1.96          & 75.96±3.34          & 87.36±8.88          & 85.6±30.46         & 9.68±21.32           \\
                                 & FP                & 79.41±3.88          & \textbf{0.4±0.37}  & {\ul 80.52±4.12}    & 83.4±1.84           & {\ul 13.34±10.71}  & {\ul 42.32±7.43}    & 85.74±1.71    & 56.57±35.33        & 32.05±28.27         & 83.37±2.56          & 38.38±27.66        & 38.38±17.39          \\
                                 & NAD               & 32.72±15.53         & 10.76±14.41        & 30.15±11.06         & 74.07±17.23         & 83.55±24.91        & 6.9±7.97            & 68.65±12.5          & 8.33±7.12          & 62.56±12.58         & 69.18±17.67         & 62.06±34.43        & 17.99±14.3           \\
                                 & CLP               & \textbf{90.71}      & 95.2               & 4.51                & \textbf{90.68}      & 91.12              & 8.03                & {\ul89.99}             & 5.51               & {\ul 80.77}         & \textbf{91.27}      & 98.53              & 1.34                 \\
                                 & FT-SAM            & 39.38±5.33          & 6.91±4.56          & 39.19±7.22          & 40.03±3.68          & \textbf{3.27±2.48} & 31.04±3.72          & 69.43±7.55          & \textbf{1.81±1.45} & 64.65±6.03          & 39.87±4.1           & \textbf{10.6±7.57} & 31.01±6.29           \\
                                 & ANP               & 84.38±1.05          & {\ul 2.38±2.68}    & \textbf{83.58±2.49} & 86.82±1.4           & 47.13±23.59        & 34.66±11.46         & 84.48±2.12          & {\ul 2.27±1.09}    & \textbf{76.71±3.06} & 86.75±1.85          & {\ul 33.79±30.94}  & {\ul 54.95±25.32}    \\ 
\multirow{-7}{*}{2}              & Ours              & {\ul 88.82±2.06}    & 13.19±6.81         & 78.97±5.71          & 85.6±2.39           & 31.51±23.92        & \textbf{46.31±14.7} & \textbf{90.32±0.68} & 47.68±31.09        & 46.77±27.62         & {\ul 89.8±2.2}      & 37.49±23.55        & \textbf{57.02±21.85} \\ [0.3ex] \hline
                                 & FT                & 84.21±3.91          & 38.24±25.64        & 53.05±19.45         & \textbf{91.8±1.28}  & 94.99±9.2          & 4.0±6.92            & {\ul 90.28±0.42}    & 2.27±0.88          & \textbf{85.21±0.72}    & 91.1±1.19     & 78.27±29.5         & 19.3±26.02           \\
                                 & FP                & 88.0±0.99           & {\ul 0.68±0.21}    & \textbf{87.88±0.87} & 88.64±0.73          & 15.0±7.01          & 50.88±4.55          & 89.77±0.87          & 65.56±32.78        & 26.5±26.56          & 88.04±0.48          & 33.86±20.47        & 46.59±12.76          \\
                                 & NAD               & 81.3±5.6            & 19.04±18.32        & 66.91±11.83         & {\ul 90.8±1.8}      & 98.83±1.93         & 1.07±1.77           & 88.97±0.83          & 3.6±1.5            & 83.48±1.46          & {\ul91.23±0.84}          & 84.72±18.6         & 13.86±17.03          \\
                                 & CLP               & \textbf{90.71}      & 95.2               & 4.51                & 90.68               & 91.12              & 8.03                & 89.99               & 5.51               & 80.77               & \textbf{91.27}      & 98.53              & 1.34                 \\
                                 & FT-SAM            & 75.21±1.44          & 2.32±0.71          & 74.41±1.67          & 81.65±0.88          & \textbf{5.37±6.77} & 52.57±5.93          & 88.46±0.55          & {\ul 1.7±0.35}     & {\ul 83.56±0.92} & 81.88±1.55          & {\ul 5.52±2.6}     & 69.23±3.15           \\
                                 & ANP               & 84.15±0.97          & \textbf{0.1±0.12}  & {\ul 86.47±1.18}    & 86.03±1.25          & 12.36±9.63         & {\ul 54.36±4.89}    & 85.22±2.51          & \textbf{0.84±0.74} & 77.85±2.0           & 86.23±1.62          & \textbf{1.19±1.3}  & \textbf{81.47±3.45}  \\
\multirow{-7}{*}{10}             & Ours              & {\ul 90.02±0.48}    & 8.72±9.69          & 83.71±7.7           & 89.01±1.52          & {\ul 8.29±10.93}   & \textbf{67.63±6.48} & \textbf{91.47±0.35} & 37.56±20.42        & 56.51±18.11         & 90.57±0.78          & 11.84±6.18         & {\ul 80.23±5.24}     \\ [0.3ex] \hline
                                 & FT                & 89.59±0.4           & 1.81±0.66          & 88.38±0.68          & \textbf{92.86±0.17} & 96.32±3.19         & 3.43±2.91           & {\ul 92.53±0.1}     & 2.41±0.37          & {\ul 86.97±0.37}    & \textbf{92.58±0.17} & 75.14±22.45        & 22.25±19.19          \\
                                 & FP                & \textbf{91.47±0.21} & {\ul 0.85±0.2}     & \textbf{90.87±0.27} & 92.23±0.17          & 14.18±4.24         & 53.23±2.22          & 92.12±0.14          & 50.47±17.2         & 34.09±10.06         & 91.66±0.22          & 32.56±9.24         & 55.17±8.1            \\
                                 & NAD               & 88.16±0.51          & 1.81±0.69          & 87.25±0.68          & {\ul 92.3±0.28}     & 88.67±6.78         & 10.34±5.98          & 92.12±0.18          & {\ul 2.32±0.35}    & 86.62±0.63          & {\ul 92.02±0.18}    & 37.5±21.7          & 55.55±18.68          \\
                                 & CLP               & 90.71               & 95.2               & 4.51                & 90.68               & 91.12              & 8.03                & 89.99               & 5.51               & 80.77               & 91.27               & 98.53              & 1.34                 \\
                                 & FT-SAM            & 90.29±0.33          & 1.24±0.33          & {\ul 89.64±0.45}    & 91.81±0.18          & 12.84±4.67         & 57.78±3.03          & \textbf{92.64±0.14} & 2.55±0.48          & \textbf{87.08±0.55} & 91.57±0.23          & 4.58±2.84          & 82.23±2.55           \\
                                 & ANP               & 84.99±1.24          & \textbf{0.02±0.02} & 88.31±1.5           & 86.62±1.61          & {\ul 3.59±1.33}    & {\ul 60.05±3.62}    & 85.64±0.99          & \textbf{0.21±0.06} & 79.1±1.78           & 86.61±1.7           & \textbf{0.25±0.14} & {\ul 85.69±1.77}     \\
\multirow{-7}{*}{100}            & Ours              & {\ul 90.91±0.56}    & 7.3±2.37           & 86.07±1.57          & 88.85±0.75          & \textbf{3.33±8.64} & \textbf{72.03±5.68} & 91.74±0.22          & 43.55±30.41        & 51.19±27.04         & 91.84±0.22          & {\ul 1.85±0.43}    & \textbf{89.98±0.38}\\  [0.3ex] \hline
\end{tabular}}
\label{tab:cifar-10-PreactResNet-18}
\end{table}
\end{landscape}

\newpage

\begin{landscape}
\begin{table}[]
\caption{The performance of the considered approaches on CIFAR-10 using VGG-19+BN. The mean and standard deviation of the performance measures over five independent runs are presented. In each setting, the best and second-best results are highlighted in bold and underlined, respectively.}
\scalebox{0.95}{
\begin{tabular}{c|c|ccc|ccc|ccc|ccc}
\hline
\multicolumn{2}{c}{Attack}                           & \multicolumn{3}{c}{BadNet}                                                  & \multicolumn{3}{c}{Blended}                                           & \multicolumn{3}{c}{BPP}                                                      & \multicolumn{3}{c}{LF}                                                          \\ \hline
SPC                              & Defense           & ACC                 & ASR                      & RA                         & ACC                 & ASR                 & RA                        & ACC                       & ASR                  & RA                        & ACC                       & ASR                     & RA                        \\ \hline
\rowcolor[HTML]{E8E8E8} 
\multicolumn{2}{c}{\cellcolor[HTML]{E8E8E8}Baseline} & \textit{90.55}      & \textit{94.22}     & \textit{5.44}        & \textit{92.08}      & \textit{99.63}     & \textit{0.34}       & \textit{89.74}      & \textit{99.28}     & \textit{0.72}       & \textit{83.57}      & \textit{13.07}     & \textit{82.38}      \\ [0.3ex] \hline
                                 & FT                & 44.44±22.79         & 23.03±33.76        & 28.25±10.01          & 66.6±29.89          & 67.64±40.58        & 8.68±11.18          & 80.02±3.63          & 4.23±3.38          & 74.09±4.92          & 33.71±15.75         & 5.79±1.98          & 33.72±15.75         \\
                                 & FP                & 82.51±5.08          & 55.73±26.33        & 37.96±21.17          & {\ul 88.16±2.84}    & 99.2±0.47          & 0.7±0.36            & {\ul 84.07±2.58}    & 6.29±4.02          & 72.77±11.32         & 77.05±10.89         & \textbf{2.31±1.46} & 78.03±10.28         \\
                                 & NAD               & 57.24±22.98         & 45.97±31.27        & 27.32±13.59          & 69.42±29.03         & 65.26±41.72        & 9.24±10.39          & 75.32±7.57          & 6.78±8.93          & 57.44±16.88         & 45.06±25.48         & 10.71±7.43         & 44.15±25.88         \\
                                 & CLP               & \textbf{88.64}      & 84.77              & 14.31                & \textbf{90.65}      & 96.37              & 3.27                & 87.63               & {\ul 1.19}         & \textbf{81.46}      & \textbf{82.83}      & 13.47              & \textbf{81.77}      \\
                                 & FT-SAM            & 56.05±12.23         & \textbf{6.56±5.06} & 53.49±10.72          & 68.95±12.01         & \textbf{4.82±4.18} & \textbf{44.54±7.84} & 79.18±5.08          & 2.68±1.47          & 73.57±6.01          & 55.59±4.18          & 3.9±3.51           & 58.28±4.63          \\
                                 & ANP               & 83.87±1.43          & {\ul 15.16±22.39}  & \textbf{72.58±18.17} & 84.17±0.87          & {\ul 36.88±29.72}  & 37.74±15.48         & 82.48±1.01          & \textbf{0.72±1.52} & {\ul 78.73±2.62}    & 77.74±1.84          & {\ul 2.38±1.29}    & 79.62±1.35          \\
\multirow{-7}{*}{2}              & Ours              & {\ul 86.59±1.41}    & 33.01±19.25        & {\ul 60.59±15.56}    & 83.53±1.87          & 45.46±13.64        & {\ul 42.24±9.73}    & \textbf{88.38±1.33} & 28.11±24.12        & 63.22±21.38         & {\ul 81.31±2.27}    & 14.12±2.29         & {\ul 80.19±2.28}    \\ [0.3ex] \hline
                                 & FT                & 80.86±3.45          & 36.96±29.04        & 51.44±21.61          & 86.17±3.15          & 72.47±22.05        & 20.75±15.49         & 87.61±0.62          & 6.25±4.17          & 75.41±6.93          & 78.12±3.47          & 2.33±0.71          & 78.87±3.0           \\
                                 & FP                & \textbf{88.62±0.74} & 76.96±13.43        & 21.45±12.34          & {\ul 90.46±0.39}    & 99.01±0.53         & 0.9±0.48            & {\ul 88.86±0.57}    & 35.23±32.96        & 43.05±25.15         & \textbf{86.18±0.58} & {\ul 2.3±0.97}     & \textbf{85.95±0.74} \\
                                 & NAD               & 84.57±4.03          & 80.76±11.9         & 17.23±10.57          & 87.96±2.82          & 91.49±10.17        & 6.97±8.46           & 86.51±1.53          & 18.38±29.23        & 47.13±25.21         & 81.1±3.69           & 4.23±2.27          & 81.48±4.1           \\
                                 & CLP               & 88.64               & 84.77              & 14.31                & \textbf{90.65}      & 96.37              & 3.27                & 87.63               & {\ul 1.19}         & {\ul 81.46}         & 82.83               & 13.47              & 81.77               \\
                                 & FT-SAM            & 84.39±3.38          & {\ul 17.75±16.93}  & 70.64±13.55          & 88.2±0.9            & {\ul 13.53±5.54}   & 56.92±3.24          & 88.33±0.64          & 8.81±15.42         & 74.95±12.28         & 82.85±1.71          & 2.57±1.2           & 83.13±1.44          \\
                                 & ANP               & 83.52±1.11          & \textbf{1.86±2.22} & \textbf{82.45±3.16}  & 84.9±0.94           & \textbf{5.81±5.11} & \textbf{59.57±4.05} & 82.77±1.22          & \textbf{0.04±0.03} & \textbf{82.42±3.0}  & 78.96±2.25          & \textbf{1.25±0.98} & 79.66±2.26          \\
\multirow{-7}{*}{10}             & Ours              & {\ul 88.54±0.33}    & 22.37±17.64        & {\ul 70.81±14.47}    & 86.15±2.27          & 24.08±15.16        & {\ul 59.37±9.03}    & \textbf{89.55±0.53} & 17.77±13.79        & 73.39±12.22         & {\ul 84.05±0.6}     & 8.62±3.75          & {\ul 83.57±1.05}    \\ [0.3ex] \hline
                                 & FT                & 88.08±0.31          & 17.09±7.96         & 74.58±6.33           & 89.51±0.39          & 79.37±18.45        & 17.46±14.86         & 90.94±0.25          & 3.33±2.16          & 83.97±2.57          & {\ul 87.17±0.31}    & 1.39±0.22          & {\ul 87.28±0.46}    \\
                                 & FP                & 88.29±0.32          & 30.3±18.7          & 63.52±16.09          & {\ul 90.78±0.43}    & 97.35±1.11         & 2.41±0.97           & {\ul 91.08±0.21}    & 36.97±23.42        & 53.81±20.24         & {\ul 87.17±0.5}     & 1.57±0.34          & 87.11±0.57          \\
                                 & NAD               & 87.1±0.51           & 15.87±12.46        & 73.54±9.38           & 88.67±0.65          & 82.6±9.2           & 15.06±7.63          & 90.54±0.27          & 2.08±0.89          & {\ul 84.07±2.92}    & 86.5±0.65           & \textbf{1.1±0.38}  & 87.02±0.61          \\
                                 & CLP               & 88.64               & 84.77              & 14.31                & 90.65               & 96.37              & 3.27                & 87.63               & {\ul 1.19}         & 81.46               & 82.83               & 13.47              & 81.77               \\
                                 & FT-SAM            & \textbf{89.97±0.26} & {\ul 6.86±4.21}    & {\ul 84.26±3.54}     & \textbf{91.01±0.19} & 16.56±7.05         & 60.14±3.78          & \textbf{91.35±0.21} & 2.94±1.47          & \textbf{85.21±2.11} & \textbf{88.98±0.14} & 1.6±0.26           & \textbf{88.83±0.25} \\
                                 & ANP               & 84.91±1.38          & \textbf{0.39±0.15} & \textbf{86.18±1.71}  & 86.14±0.91          & \textbf{1.85±1.47} & {\ul 65.17±1.56}    & 81.64±0.52          & \textbf{0.01±0.01} & 82.48±2.13          & 80.17±2.35          & {\ul 1.24±0.66}    & 80.38±2.72          \\
\multirow{-7}{*}{100}            & Ours              & {\ul 88.69±0.25}    & 26.22±16.89        & 67.75±13.87          & 83.84±3.5           & {\ul 2.29±1.09}    & \textbf{73.14±2.49} & 89.64±0.63          & 22.04±26.73        & 68.89±22.98         & 82.52±3.2           & 6.45±3.85          & 82.77±3.57  \\ [0.3ex] \hline                  
\end{tabular}}
\label{tab:cifar-10-VGG19+BN}
\end{table}

\end{landscape}

\end{document}